\documentclass[runningheads]{llncs}
\usepackage{graphicx}
\usepackage{booktabs}
\usepackage{amsmath}
\begin{document}
\title{Inter-case Predictive Process Monitoring: A candidate for Quantum Machine Learning?}
\titlerunning{Quantum Machine Learning for Predictive Process Monitoring}
%
\author{Stefan Hill\inst{1,2} \and
David Fitzek\inst{2,3} \and
Patrick Delfmann\inst{1} \and
Carl Corea\inst{1}}
\authorrunning{S. Hill et al.}
%
\institute{University of Koblenz, Research Group Process Science, 56070 Koblenz, Germany
\email{\{shill,ccorea,delfmann\}@uni-koblenz.de} \and
Chalmers University of Technology, Department of Mircotechnology and Nanoscience, 412 58 Gothenburg, Sweden \\
\email{\{stefanhi,davidfi\}@chalmers.se} \and
Volvo Group Trucks Technology, 405 08 Gothenburg, Sweden}
\maketitle              
\begin{abstract}
Regardless of the domain, forecasting the future behaviour of a running process instance is a question of interest for decision makers, especially when multiple instances interact. Fostered by the recent advances in machine learning research, several methods have been proposed to predict the next activity, outcome or remaining time of a process automatically. Still, building a model with high predictive power requires both – intrinsic knowledge of how to extract meaningful features from the event log data and a model that captures complex patterns in data. This work builds upon the recent progress in inter-case Predictive Process Monitoring (PPM) and comprehensively benchmarks the impact of inter-case features on prediction accuracy. Moreover, it includes quantum machine learning models, which are expected to provide an advantage over classical models with a scaling amount of feature dimensions. The evaluation on real-world training data from the BPI challenge shows that the inter-case features provide a significant boost by more than 4\% in accuracy and quantum algorithms are indeed competitive in a handful of feature configurations. Yet, as quantum hardware is still in its early stages of development, this paper critically discusses these findings in the light of runtime, noise and the risk to overfit on the training data. Finally, the implementation of an open-source plugin demonstrates the technical feasibility to connect a state-of-the-art workflow engine such as Camunda to an IBM quantum computing cloud service.

\end{abstract}

\section{Introduction}


Today, enterprises, universities or public institutions track large amounts of time-stamped data stored in event logs. Propelled by the ongoing advances in machine learning (ML) research and based on the domain knowledge of business process management (BPM), Predictive Process Monitoring (PPM) is a collection of techniques that aim to predict the future behaviour of business processes using historical data as features~\cite{maggi2014predictive-process-monitoring}. As business processes in applications such as logistics, airport operation or hospital management, are based on complex interactions, the outcome of a single process instance also depends on other concurrently executing instances and their states~\cite{difrancescomarino2022ppm}. Therefore, several attempts have been made to incorporate the inter-case dependencies~\cite{denisov2019performance-spectrum,grinvald2021inter-case-variants,pourbafrani2022inter-case-remaining-time,senderovich2017intra-inter}.

Based on theoretical computer science and quantum physics, the idea of using quantum systems for computation has been developed since 1982, when Feynman published the idea of 'simulating quantum physics with quantum physics'~\cite{feynman1982quantum-physics}. What Feynman means is to use real quantum systems such as electrons or photons to perform calculations. Since then, it has been shown that quantum computers can solve combinatorial problems in complexity classes beyond the classical ones. The most prominent example is Shor's algorithm, which efficiently derives the prime factorisation for large composite integers exponentially faster than classical implementations~\cite{shor1994primes}. Apart from this algorithm, which has the potential to revolutionise cryptography, applications of quantum algorithms range from chemistry~\cite{Kandala2017HardwareefficientVariationalQuantum,Yamamoto2019naturalgradientvariational} to finance~\cite{Egger2020QuantumComputingFinance,Emmanoulopoulos2022QuantumMachineLearning,Pistoia2021QuantumMachineLearning} to industrial optimisation~\cite{Streif2021Beatingclassicalheuristics}. In short, the advantage of using a quantum computer lies in its higher dimensional vector space to perform calculations.

The novelty presented in this paper is to solve the inter-case PPM problem with a variational quantum algorithm and the implicit quantum kernel estimation on a support vector machine~\cite{havlicek2019quantum-feature-space,schuld2020circuit-centric}, which - to the best of our knowledge - has not been tested ever before. In our evaluation, we can empirically show that the quantum kernels were able to inherently capture the PPM-specific patterns between cases, thus significantly outperforming classical kernel-based approaches in terms of accuracy. Since training time is an important decision factor in production scenarios, we also tested stratification sampling and were able to show that training time can be reduced by 85\% while maintaining the same accuracy by considering only 50\% of the features. All prototypical implementations can be found in an open source library\footnote{\url{https://gitlab.com/stefanhill/qppm}} and follow standard industrial library interfaces, so cross-validation and model selection methods are supported and the classifiers work with all common open-source Python-based frameworks.

Even though recent publications might draw an overly optimistic picture of the ongoing hype around quantum technology and the European Union equipped the initiative Quantum Flagship with an enormous budget of 1 billion euros~\footnote{\url{https://qt.eu/about-quantum-flagship/introduction-to-the-quantum-flagship/}}, we would like to keep the expectations on a realistic level. Quantum systems in their current dimensions are called noisy intermediate-scale quantum (NISQ) devices and are still in their early stages of development~\cite{preskill2018nisq}. To give an example, \cite{arute2019google-sycamore} claim to be able to solve a problem in 200 seconds on Google's 53 qubits processor Sycamore, which would take at least 10,000 years to be solved on a classical computer while a Chinese research group at the University in Hefei reports achieving first promising results on a photon-based QPU with 76 qubits ~\cite{zhong2020quantum-supremacy}. Yet, the most advanced systems like the IBM Eagle QPU consist of 127 noisy qubits, and a 433 qubit processor was released in October 2022~\footnote{\url{https://www.ibm.com/quantum/roadmap}} which is more than enough to run the algorithms that we are using in our experiments. Thus, to demonstrate the technical feasibility of running the algorithms on IBM hardware, we additionally implemented an open-source plugin that connects the workflow engine Camunda to the IBMQ cloud service. By establishing this link between emergent scientific fore-front prototypes and solution-oriented mature business tools in an industrial-grade environment, we argue that the combination of quantum computing and BPM serves as an ideal draught horse for fundamental research in both areas.

This is why the paper is structured as follows. First, we introduce the inter-case PPM problem and we will give a brief introduction to quantum computing. Second, we explain the pipeline for our experiments that includes both the feature extraction from the event log and the quantum kernel classifier. Next, we present the improvements in accuracy and runtime for our selection of classifiers. In light of our findings, we discuss how they contribute to the current landscape of research in quantum kernel methods. We conclude by showing directions for research at the intersection of quantum computing and BPM and present a demonstration of the prototype.

\section{Background}
In this section, we will briefly explain the terminology of the inter-case PPM problem followed by an introduction to quantum computing. Furthermore, we will sketch how quantum circuits are plugged into common ML models, to build the so-called hybrid quantum-classical ML models for our experiments.

\begin{figure*}[ht]
	\centering
	\includegraphics[width=0.9\textwidth]{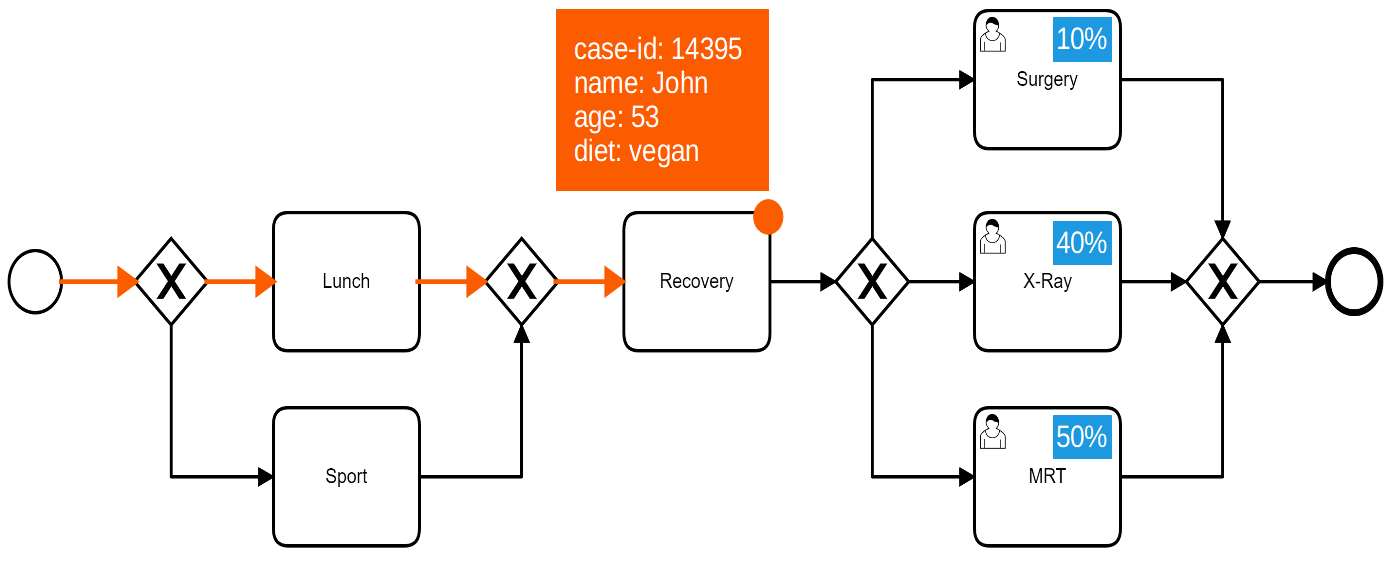}
	\caption{Example process with running instance (orange dot). An exemplary predictive process monitoring problem is the prediction of the next task (blue) based on historical events (orange arrows) and case attributes (orange box).}
	\label{fig:example-ppm}
\end{figure*}

\subsection{Predictive Process Monitoring}
Processes can be extracted and stored in a so-called \textit{event log}. Such an event log consists of a set of \textit{traces} that refer to a business \textit{case} which consists of an arbitrary number of distinct \textit{events}.. In practice, machine-readable formats such as the eXtensible Event Stream (XES) XML-based standard are used\footnote{\url{http://www.xes-standard.org/}}.

Consider for example the process model of a hospital as in Figure~\ref{fig:example-ppm}. In that scenario, the patient would be assigned a case and takes part in several activities which will be logged as \textit{trace} \(\sigma_x\) of events \(e \in \mathcal{E}\), where \(\mathcal{E}\) is the universe of all events. The first event of a new patient could be lunch. This event would carry some information like the \textit{case id}, the \textit{activity} with its corresponding \textit{activity name} (standard concept:name) and the \textit{timestamp}. These three main properties of the event are usually called the \textit{control flow}. An exhaustive formal description of the event log can be found in the appendix and in~\cite{teinemaa2019diss}. This allows us to formulate the problem of predictive process monitoring~\cite{senderovich2017intra-inter}.


\begin{definition}[Predictive Process Monitoring (PPM)]
	Given a (possibly running) case \(\sigma_x\), the predictive process monitoring problem (PPM) is to find a function \(f: \mathcal{E}^* \rightarrow \mathcal{Y}\) that accurately maps \(\sigma_x\) to the corresponding label \(y(\sigma_x)\).
\end{definition}

The label refers to anything that is called a prediction type, e.g. the next activity or remaining time a business case until completion~\cite{difrancescomarino2022ppm}. In that manner, the PPM problem becomes a ~\textit{supervised learning task} (SLT)~\cite{senderovich2017intra-inter}. 

Casting the PPM problem into a SLT requires a function \(g\) which transforms sub-traces (also called prefixes) from an event log to features for the ML algorithm. Moreover, a prerequisite to solve the SLT is that observations are independent identically distributed (i.i.d.) feature-outcome pairs \((x_i, y_i)\). However, the event data in a prefix log \(L^*\) is highly correlated for prefixes of the same case (\emph{intra}-case dependency) and for every two prefixes that run at the same time or share limited resources in the same environment (\emph{inter}-case dependency). \cite{senderovich2017intra-inter} define the Sequence-To-feature Encoding Problem (STEP) as follows. 

\begin{definition}[Sequence-To-feature Encoding Problem (STEP)]
	Let \(\mathcal{L}^*\) be an extended event log that contains all prefixes of the sequences in \(L\). Solving the STEP problem is to find a function \(g: \mathcal{E}^* \times \mathcal{Y} \times 2^{L} \rightarrow \mathcal{X} \times \mathcal{Y}\) such that the result of its operation, \(\{g(\sigma_i, y(\sigma_i), L^*)\} \subset \mathcal{X} \times \mathcal{Y}\), is an i.id. sample of feature-outcome from \(\mathcal{X} \times \mathcal{Y}\).
\end{definition}

Recently, a number of STEP encodings for the function \(g\) have prevailed~\cite{teinemaa2019benchmark-outcome-oriented}. The most simple ones are the \textit{static} and the \textit{last state} encoding, which only take the attributes of an instance and the last state into account. These two encodings do not include historical information about the business case. \textit{Aggregation-based} encodings count the number of occurences of an activity or indicate by a boolean whether the activity was included in the past execution of an instance~\cite{leontjeva2015complex}. While the aggregation-based encodings include the number of past events, they still do not contain the development of variables over the history of the case. This is where the \textit{index-based} encoding comes into play. It appends all states into a feature vector, thus, is lossless by design. To ensure that the feature vectors are of same length, missing values are padded with a default value.

The challenge in PPM is to combine sequential and contextual case information. Hence, the encoding must reflect the order of events, as well as dynamic and static case attributes assigned to the case. As cases that are executed at the same time or share resources are highly inter-related, it is important to encode this information as well. Some approaches incorporate the inter-case dependencies into the encoding by appending aggregated features~\cite{grinvald2021inter-case-variants,senderovich2017intra-inter,senderovich2019inter-case-encoding}, or capture the temporal dynamics by grouping instances in batches~\cite{klijn2020performance-spectrum,pourbafrani2022inter-case-remaining-time}.

\vspace{-.4cm}
\begin{table}[ht]
    \centering
    \caption{Inter-case features of our experiments.}
    \begin{tabular}{|p{0.18\textwidth}|p{0.80\textwidth}|}
        \hline
        \textbf{peer\_cases} & Counts the number of concurrently executed cases in the time window. \\
        \hline
        \textbf{peer\_act} & Counts the total number of triggered activities of all cases during the time window. \\
        \hline
        \textbf{res\_count} & Counts the number of resources working in the time window. \\
        \hline
        \textbf{avg\_delay} & Calculates an average delay from the activity transitions in the time window relatively to the average transition times of the training log. \\
        \hline
        \textbf{freq\_act} & Returns the topmost frequent activity in the time window. \\
        \hline
        \textbf{top\_res} & Returns the topmost frequently used resource in the time window.\\
        \hline
        \textbf{batch} & Returns an indicator of how prone the upcoming possible events are to end up in a batch. A batch, here, forms when there is a task such as taxes that is only performed once in a cycle, thus, all instances have to wait until a specific date irrespective of their arrival time at their current activity.\\
        \hline
    \end{tabular}
    \label{tab:inter-case-features}
\end{table}
\vspace{-.4cm}
We build our features based on the central concept introduced by~\cite{grinvald2021inter-case-variants}, which is to span a time window that moves alongside the current event to be encoded. All events inside the window are called \textit{peer events} and the respective cases are defined to be \textit{peer cases} of the current running instance. Using this notion of peer cases, we delineate the set of inter-case features in Table~\ref{tab:inter-case-features} based on suggestions by~\cite{grinvald2021inter-case-variants} and~\cite{senderovich2019inter-case-encoding}. In contrast to the intra-case features such as the aggregation or the index-based encoding, these inter-case features do not encode information of a single sub-trace. They rather aggregate metrics that depend on the concurrently executed process instances. Now, an inter-case feature is defined as STEP solution where the feature space \(\mathcal{X}\) is replaced a bi-dimensional space \(\mathcal{X}_1 \times \mathcal{X}_2\). Here, \(\mathcal{X}_1\) represents the intra-case part of the solution i.e. the event sequence and \(\mathcal{X}_2\) consists of the inter-case features, which are aggregated relative to the last event. An example for such a feature vector could be the last four activities of the patient in the hospital plus the total number of doctors (resources) and the total number of other patients in the hospital during the last two hours. Whether the patient proceeds with an operation or has to be scheduled for lunch depends on his medical record as well as on the staffing of the station and the load. A more detailed introduction to how inter-case features are assembled can be found in the supplementary materials~\footnote{Supplementary material A:~\url{https://bit.ly/qppm-supplementary}}.



\subsection{Quantum Computing}
Quantum computing is a fundamentally different approach of computation, which takes advantage of manipulating properties of quantum mechanical systems. The constitutional difference between classical and quantum computation is that the quantum bit is represented as a two-dimensional tensor whereas a classical bit lives in one dimension only. This circumstance implies an exponential increase of the state space. Instead of using high and low voltage on a transistor to represent 0 and 1 in a classical bit, quantum bits are represented by utilizing quantum properties like the spin states of an electron or the polarization states of a photon~\cite{hey1999introduction-qc}. Mathematically the state of these two-level quantum systems with a ground state \(\vert 0 \rangle \) and an excited state \(\vert 1 \rangle \) is denoted by the state vector:
\begin{equation}
	\vert \psi \rangle = \alpha \vert 0 \rangle + \beta \vert 1 \rangle = \begin{pmatrix}
		\alpha \\
		\beta
	\end{pmatrix}
\end{equation}
where \(\alpha\) and \(\beta\) are complex numbers satisfying \(\vert \alpha \vert ^2 + \vert \beta \vert ^2 = 1\). In this matter, unlike a classical bit, the qubit has infinitely many possible states - all superpositions of \(\vert 0 \rangle \) and \(\vert 1 \rangle \). A measurement on the qubit state using basis \(\vert 0 \rangle \) and \(\vert 1 \rangle \) will return 0 with probability \(\vert \alpha \vert ^2\) and 1 with probability \(\vert \beta \vert ^2\)~\cite{ferrini2021ln-qc}. Here, the Bloch sphere is used as a tool to represent the state of a qubit. An arbitrary superposition of a qubit \(\vert \psi \rangle \) can be written as \(\vert \psi \rangle = cos \frac{\phi}{2} \vert 0 \rangle + e^{i \theta} sin \frac{\phi}{2} \vert 1 \rangle \).

If there are \(N\) qubits in a system, the total state of that system can be a superposition of \(2^N\) different states: \(\vert 000...00 \rangle, \vert 100...00\rangle, ... \vert 111...11 \rangle \). A classical system can be in one of these \(2^N\) states but not in a superposition of several of them. The superposition of a qubit can be enforced by applying a Hadamard gate. Likewise classical gates such as an AND or an OR gate, quantum gates manipulate the state of one or multiple qubits. A brief introduction into quantum gates can be found in the supplementary materials~\footnote{Supplementary material B:~\url{https://bit.ly/qppm-supplementary}}.
\begin{figure}[ht]
	\centering
    \includegraphics[width=0.35\textwidth]{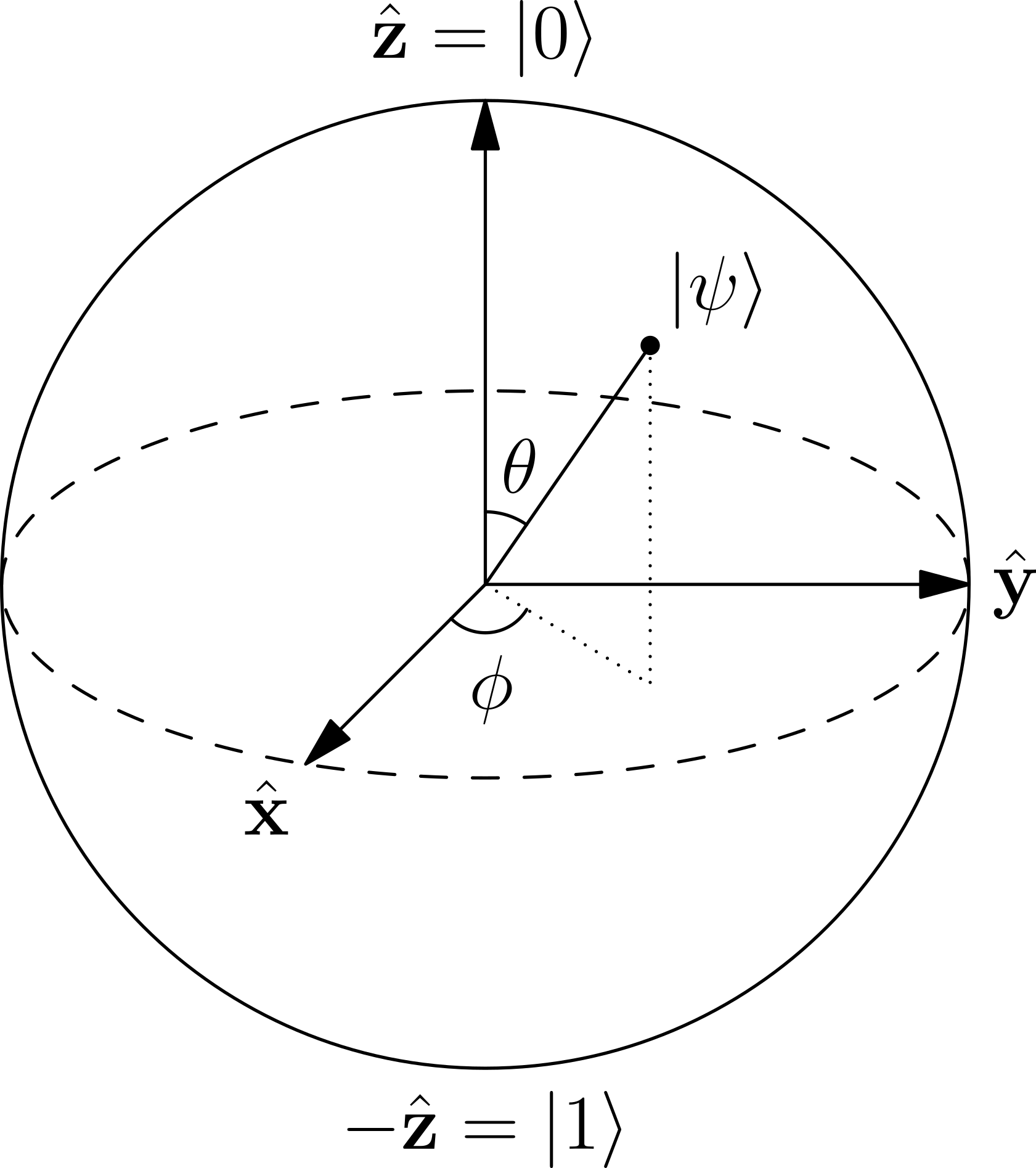}
    \caption{Bloch sphere to visualise the states of a qubit. States \(\hat{z} = \vert 0 \rangle \) and \(-\hat{z}=\vert 1 \rangle \) are the most common measurement bases where \(\hat{z}\) is the ground state with zero energy~\cite{glosser2022bloch-sphere}.}
	\label{fig:blochsphere}
\end{figure}

\subsection{Quantum Kernel Methods}
The idea behind quantum kernel methods is to take advantage of the higher dimensional Hilbert space when calculating the inner products for the kernel trick~\cite{scholkopf2001representer,scholkopf2002kernel}. It can be shown that the embedding of a data vector \(\mathbf{x}\) into a quantum state \(\vert \phi(\mathbf{x}) \rangle \) fulfils the definition of a feature map \( \phi: \mathcal{X} \rightarrow \mathcal{F} \) with \(\mathcal{F}\) being a quantum feature space~\cite{schuld2019feature-space-extended}. Instead of using the notion of an inner product, a linear model in the quantum representing kernel Hilbert space (RKHS) is defined in Dirac notation:
\begin{equation}
	f(x; w) = \langle w \vert \phi(x) \rangle
\end{equation}
with \(\vert w \rangle \in \mathcal{F}\) being a weight vector living in the feature space. The concept of interpreting \(x \rightarrow \vert \phi(x) \rangle \) as feature map opens up all possibilities of classical kernel methods for the quantum world. The backbone is a \textit{feature-embedding} circuit \(U_{\phi}(x)\) that acts on a ground state \(\vert 0...0 \rangle \) of a Hilbert space \(\mathcal{F}\) as \(U_{\phi}(x)\vert 0...0 \rangle \)~\cite{schuld2019feature-space-extended}.

The real benefit of a quantum kernel is hidden in non-Gaussian feature maps that are not easy to simulate classically. \cite{havlicek2019quantum-feature-space} propose such a \(n\)-qubits feature map based on Hadamard and z-gates, called~\textit{zz feature map}:
\begin{equation}
	\mathcal{U}_{\phi}(\mathbf{x}) = U_{\phi}(\mathbf{x}) H^{\otimes n} U_{\phi}(\mathbf{x}) H^{\otimes n}
\end{equation}
with
\begin{equation}
U_{\phi} (\mathbf{x}) = exp \left( i \underset{S \subseteq [n]}{\sum} \phi_S(\mathbf{x}) \underset{i \in S}{\prod} Z_i \right)
\end{equation}

\begin{figure}[h]
	\centering
	\includegraphics[width=0.6\textwidth]{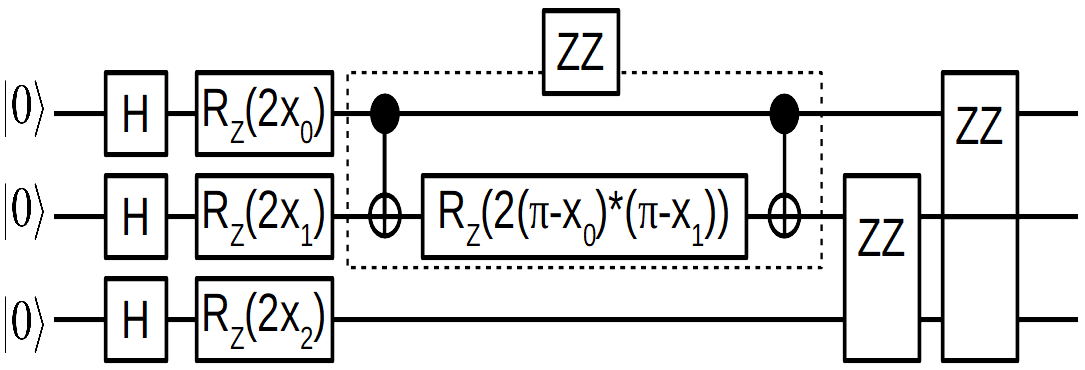}
	\caption{Feature map based on rotations along the z-axis and Hadamard gates as proposed by~\cite{havlicek2019quantum-feature-space} on an exemplary circuit with 3 qubits.}
	\label{fig:zz-feature-map}
\end{figure}

To gain a better understanding of how the feature map looks like, Figure~\ref{fig:zz-feature-map} visualizes the underlying circuit. Qubits are prepared by the Hadamard and rotational z-gates and entangled in the section with zz-gates. The circuit is also implemented in the evaluation framework. Additionally, we were using an adapted variant of the circuit that encodes on the y-axis instead of the z-axis.

While implementing the feature map is only the first step to embed a data vector \(\mathbf{x}\) into the circuit, the second one is to integrate the kernel into a classification model. Here, the two most prominent approaches were developed independently by~\cite{schuld2019feature-space-extended} and~\cite{havlicek2019quantum-feature-space}. Both articles show that it is feasible to either implement the kernel explicitly as \textit{variational quantum classifier} (VQC) or to build a hybrid model with a circuit called \textit{quantum kernel estimator} (QKE) which is plugged into a classical SVM and calculates the kernel matrix implicitly~\cite{vapnik1995svm}.

In the implicit approach, the quantum computer is used to estimate the inner products \(\kappa(x, x') = \langle \phi(x), \phi(x') \rangle \) for a kernel-dependent model. Thus, the quantum computer is required to implement the state preparation routine \(\mathcal{U_{\phi}}(x)\) for any \(x \in \mathcal{X}\)~\cite{schuld2019feature-space-extended}. The decision boundary is determined classically using a SVM. Formally, the kernel circuit is described by:
\begin{equation}
	\kappa(x, x') = \langle 0...0 \vert V_{\phi}(\mathbf{x}) V_{\phi}^{\dagger}(\mathbf{x'}) \vert 0...0 \rangle
\end{equation}
where the function \(V_{\phi}\) is implemented twice, once to encode the training sample \(\mathbf{x}\) and once in its adjoint form \(V_{\phi}^{\dagger}\) to encode the sample \(\mathbf{x'}\) to which we want to measure the distance. The idea behind the variational classifier is to perform the classification explicitly in the quantum Hilbert space. Similar to a perceptron, the algorithm relies on a parametrized weight matrix, which is updated by an optimizer after determining the classification error. Formally, there exists a weight matrix \(W = W(\theta)\) which learns a model \(\vert w(\theta) \rangle = W(\theta) \vert 0...0 \rangle \) that is evaluated by the linear model function \(f(x; w) = \langle w \vert \phi(x) \rangle\). The simplest circuit with one weight layer is given by \(W(\theta)U_{\phi} \vert 0...0 \rangle \)~\cite{schuld2019feature-space-extended}. ~\cite{schuld2020encoding} also introduce circuits with multiple layers and find that a higher number of parameterized weight matrices increases the model capabilities to approximate an arbitrary function.

\section{Experimental setup}
The goal of our experiments is threefold. First, we aim to compare quantum and classical methods to solve the inter-case PPM problem. Second, we want to investigate properties of the quantum kernels. Third, we want to build our test framework as reusable as possible which allows us to integrate it into existing software. Thus, the pipeline for our experiment is straightforward: Intra-case feature vectors are built from the event log augmented by the inter-case features as in Table~\ref{tab:inter-case-features}. Next, a selection of ML models is trained. Features are encoded on the quantum circuit, if applicable to the algorithm. Forthwith, we will briefly justify the choice of our training data and state the configuration of our simulator.

\vspace{-.4cm}
\begin{table}[ht]
	\centering
	\caption{Statistics about the raw event logs and the version preprocessed by variants.}
	\begin{tabular}{cccccc}
		\toprule
		\rule[-1ex]{0pt}{2.5ex} \textbf{Dataset} & \textbf{no. cases} & \textbf{no. events} & \textbf{no. activities} & \textbf{no. variants} & \textbf{median case time} \\
		\midrule
		\rule[-1ex]{0pt}{2.5ex} BPIC17 & 31,509 & 1,160,405 & 26 & 15484 & 19.1d \\
		\rule[-1ex]{0pt}{2.5ex} BPIC17-P & 18,112 & 507,161 & 25 & 2087 & 20.4d \\
		\rule[-1ex]{0pt}{2.5ex} RTFM & 150,370 & 561,470 & 11 & 231 & 28.8w \\
		\rule[-1ex]{0pt}{2.5ex} RTFM-P & 150,270 & 560,551 & 11 & 131 & 28.3w \\
		\bottomrule
	\end{tabular}
	\label{tab:dataset-stats}
\end{table}
\vspace{-.4cm}

A well-known source for evaluation datasets in process mining is 4TU Center for Research Data\footnote{\url{https://data.4tu.nl/}} which provides event logs for the Business Process Intelligence (BPI) Challenge. Therefore, the following two logs will be used as raw data for the evaluation (see Table~\ref{tab:dataset-stats} for some general statistics about the dataset).

\textbf{Loan Application (BPIC17)} This log comprises loan applications of a Dutch financial institute. It is selected for the evaluation as it contains information about resources working on the tasks. 

\textbf{Road Traffic Fine Management (RTFM)} The log is extracted from an Italian police office and contains all events related to the fine collection process. It is included in the evaluation because there are inherent inter-case dependencies. For example, unpaid fines are collected once a year, which means that all cases of the same year depend on this event.

Because the quantum algorithms will be simulated on a classical computer, the datasets must have been reduced to conduct tests in time. Hence, the number of cases will be reduced by first removing all variants that occur only once. The RTFM log does not seem to be affected that much by the filtering as most of the cases depict mainstream behaviour. Also, the BPIC17 dataset still contains more than 500.000 events which will be too large for the quantum kernel simulation. Thus, secondly, all datasets will be shortened by picking a random timeframe.

\vspace{-.4cm}
\begin{table}[ht]
	\centering
	\caption{Statistics about the datasets with reduced sample sizes.}
	\begin{tabular}{cccccc}
		\toprule
		\rule[-1ex]{0pt}{2.5ex} \textbf{Dataset} & \textbf{no. cases} & \textbf{no. events} & \textbf{no. activities} & \textbf{no. variants} & \textbf{date range} \\
		\midrule
		\rule[-1ex]{0pt}{2.5ex} BPIC17-S & 1,940 & 57,107 & 24 & 731 & 20160215-20160415 \\
		\rule[-1ex]{0pt}{2.5ex} RTFM-S & 3,321 & 12,406 & 11 & 27 & 20030501-20040430 \\
		\bottomrule
	\end{tabular}
	\label{tab:train-val-dataset-stats}
\end{table}
\vspace{-.4cm}

This procedure is similar to what one would expect in a real-world setting: Often classifiers are trained on historical data of a certain timeframe and used for predictions in the following period. To ensure there are enough cases inside the time window, a period of around two to three times the preprocessed median case time will be chosen. The resulting global statistics about the remaining datasets are shown in Table~\ref{tab:train-val-dataset-stats} in the appendix. Furthermore, all experiments will be conducted on cross-validation with three folds which is a typical value used in other benchmarks for PPM~\cite{rama-maneiro2022deep-learning-benchmark,teinemaa2019benchmark-outcome-oriented}.

The experiments are conducted on a classical computer that simulates the quantum circuits using the \texttt{default\_qubit} simulator\footnote{Although we are using the simulator, interfacing to an actual quantum computer is possible with \texttt{pennylane}. In our implementation, one can define a connection to a cloud-hosted machine from~\url{https://quantum-computing.ibm.com/}} from \texttt{pennylane} with 1000 measurement shots\footnote{\url{https://bit.ly/3WB0Nb8}}. Circuits are accelerated using \texttt{jaxlib} which requires the operating system to be Linux-based. Experiments run on a server with Ubuntu 18.04.6 LTS and Python 3.8.13 installed\footnote{A list of libraries and dependencies can be found in the \texttt{requirements.txt} file}. The CPU is an Intel i9-9820X with 10 cores and 20 threads on 4.20 GHz and there is 64 Gb of RAM.

\section{Evaluation}

In our experiments, we focus on three aspects. First, we measure the influence of adding inter-case features. Next, we investigate to which extent quantum kernel methods improve the prediction's accuracy compared to classical kernels. And third, we present ways to accelerate the training for quantum kernels against the backdrop that our experiments are still simulated and real hardware currently runs on a limited number of qubits. For reference, we also include common classification models from PPM and the VQC into our pipeline.

\begin{figure}
    \centering
    \includegraphics[width=\textwidth]{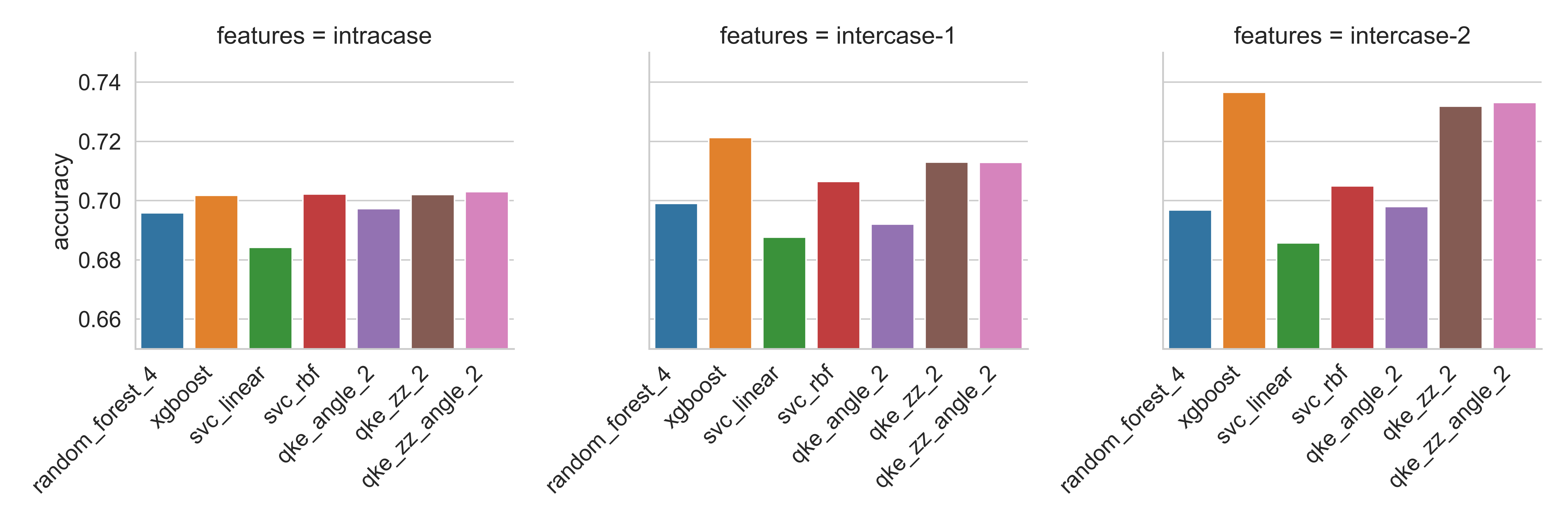}
    \caption{Increase in accuracy when adding inter-case features. The quantum kernels (brown and pink) lead to a more extreme increase than the classical kernel (red) and compete with XGBoost (orange).}
    \label{fig:all-accuracies-graph}
\end{figure}

First, we run a grid search with to find the length for the intra-case feature which saturated for index-based encodings with more than four steps. Consequently, a four-step index-based encoding forms the baseline for inter-case scenarios (see left graph in Figure~\ref{fig:all-accuracies-graph}). In the following, all intra-case features are augmented with exactly one inter-case feature. As all seven inter-case encoders share the hyperparameter of the peer cases window, we try three configurations -- 0.15, 0.30, and 0.5 -- and multiply by 20.4 weeks, which is the median case duration on the RTFM dataset.~\cite{grinvald2021inter-case-variants} choose the window to be 20\% or sometimes even less of the average case length on other datasets, but do not explain in detail which criteria to take into account when determining the window size. The parameter has an influence on the information gain of the features. If set too small, no dependencies were captured. If set too large, dependencies were taken into account that would not have existed in the real world.

As there are plenty of encoding combinations to be simulated, we decided to conduct the tests on the RTFM-S dataset, which is the optimal trade-off for the number of training instances. On the BPIC17-S dataset, it would not be possible to conduct tests in bearable time. The tree-based algorithms will be random forests with tree depths 3 and 4 as well as the standard XGBoost algorithm. We chose these hyperparameters because we found them to be ideal in previous experiments. Moreover, the linear and radial base (RBF) kernel of the classical SVM will be evaluated against the angle, zz and the angle\_zz feature map with one and two kernel layers. Also here, we chose the zz feature map, because it delivered competitive results in a quick intra-case evaluation and is an upcoming standard kernel in the quantum community~\cite{havlicek2019quantum-feature-space}. We include variational quantum circuits and try several number of layers as this hyperparameter is expected to be the most influential~\cite{schuld2020circuit-centric}.

The results of the benchmark can be found in Table~\ref{tab:inter-case-acc-rtfm} and Figure~\ref{fig:all-accuracies-graph}. Noteworthy, the quantum kernel with a zz feature map delivers throughout higher accuracies than the SVM with RBF kernel. Still, one can see that XGBoost achieves high accuracies for all features. To mention the importance of the inter-case features in comparison to intra-case features only, for XGBoost, the increase is about 3.6\% and for the quantum kernel accuracies are about 1.7\% higher than on an \textit{index\_bsd\_4} baseline encoding. The VQC delivers comparably low accuracies -- it might be possible that the optimizer found a local, but not a global minimum of the error function.

\begin{table}        
	\centering
	\caption{Accuracies on the RTFM-S event log. Values are averaged for three configurations of the inter-case feature extraction. Best classical and quantum classifier in bold.}
	\begin{tabular}{lrrrrrrr}
    
		\toprule
		{} &  peer\_cases &  peer\_act &  res\_count &  avg\_delay &  freq\_act &  top\_res &   batch \\
		\midrule
		rf\_4 &      0.7031 &    0.7026 &     0.6945 &     0.7051 &    0.6952 &   0.6971 &  0.6960 \\
		\textbf{xgboost}         &      \textbf{0.7384} &    \textbf{0.7351} &     \textbf{0.7247} &     \textbf{0.7298} &    \textbf{0.7157} &   \textbf{0.7036} &  \textbf{0.7018} \\
		svc\_linear      &      0.6880 &    0.6859 &     0.6848 &     0.6906 &    0.6951 &   0.6850 &  0.6843 \\
		svc\_rbf         &      0.7063 &    0.7080 &     0.7073 &     0.7076 &    0.7106 &   0.7031 &  0.7023 \\
		vqc\_zz\_2        &      0.0816 &    0.0672 &     0.1159 &     0.1355 &    0.1233 &   0.1237 &  0.1257 \\
		qke\_angle\_2     &      0.6917 &    0.6903 &     0.6923 &     0.6942 &    0.7011 &   0.6876 &  0.6875 \\
		\textbf{qke\_zz\_2}        &      \textbf{0.7175} &   \textbf{0.7270} &     \textbf{0.7108} &     \textbf{0.7196} &    \textbf{0.7113} &   \textbf{0.7029} &  \textbf{0.7021} \\
		qke\_zz\_a\_2  &      0.7184 &    0.7274 &     0.7097 &     0.7190 &    0.7112 &   0.7027 &  0.7021 \\
		\bottomrule
	\end{tabular}
	\label{tab:inter-case-acc-rtfm}
\end{table}

To further boost accuracy, we test a combination of multiple inter-case features on the RTFM-S dataset. As the \textit{top\_res} and the \textit{batch} encoding did not provide any advantage in the previous inter-case experiment they are neglected. The time window parameter is deliberately set to 10.2 weeks which is 50\% of the median case time. In that way, a total number of ten combinations is tested on each classifier. As the dataset is still relatively small (12,406 samples), we do not test more than two inter-case features to avoid overfitting. For further details about overfitting and the exact values for the accuracies, we refer to the regression analysis and the tables in the respective supplementary materials~\footnote{Supplementary material D+E:~\url{https://bit.ly/qppm-supplementary}}.

While XGBoost was still far ahead with only one inter-case feature, this is no longer the case with two inter-case features. On the one hand, XGBoost delivers the highest overall score (74.55\%) on the feature \textit{peer\_cases+freq\_act} while 72.90\% is the highest score for the quantum kernel with zz feature map on the same encoding. On the other hand, quantum algorithms achieve a high accuracy on the encoding \textit{peer\_act+avg\_delay} (74.51\%) where the maximum score for XGBoost is 73.51\% and 70.33\% for the RBF kernel. As five out of ten results are higher for a quantum kernel than XGBoost and all results of a zz feature map outperform RBF, we conclude that quantum kernels are a competitive choice for the inter-case PPM problem.

\begin{figure*}[ht]
\centering
	\includegraphics[width=\textwidth]{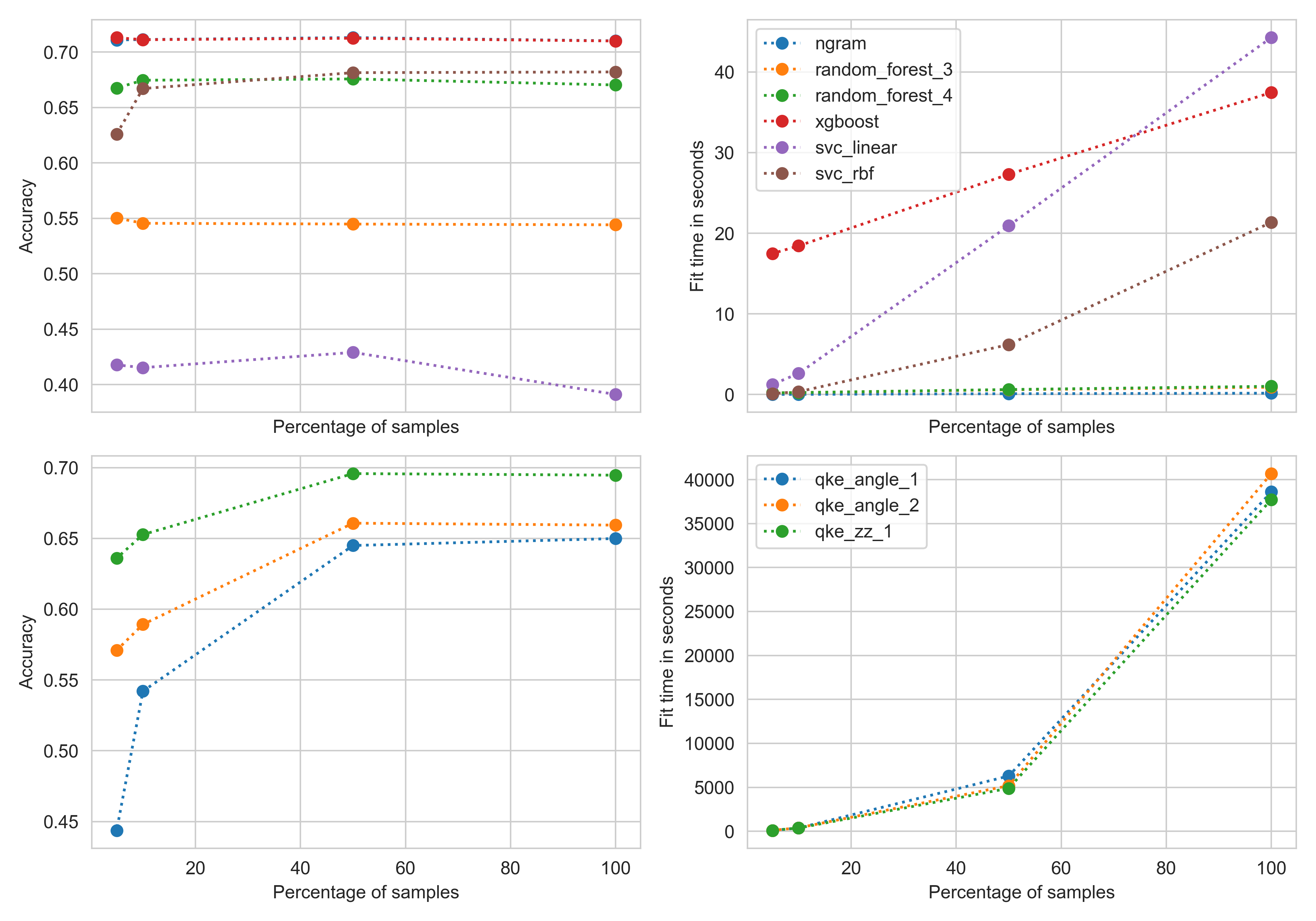}
	\caption{Accuracies and fit times for a selection of algorithms on the BPIC17-S dataset with 57,107 samples. Quantum algorithms in the lower part of the figure, classical algorithms in the upper part.}
	\label{fig:samples-acc-fit-time}
\end{figure*}

Because the bottleneck of quantum simulations is the training time, it is interesting to find out how to reduce the number of kernel evaluations. The most simple approach is in resampling the training dataset and using only a certain percentage of the training features. Figure~\ref{fig:samples-acc-fit-time} shows accuracies and fit times for quantum and classical methods on the BPIC17-S dataset when we applied stratification sampling. There is no decrease in accuracy for all selected algorithms when only 50\% percent of the training samples are taken into account. However, runtime decreases from 40683 seconds to 4886 seconds for the \textit{qke\_angle\_2} which is less than 15\% of the original train time.


\section{Directions}

When taking a closer look to research in QML, there are mainly two approaches to show a quantum advantage of specific algorithms. First, by proving that certain advantages exist theoretically due to the structure of a problem. Second, by investigating the quantum advantage in an exploratory way and simply apply quantum algorithms on any kind of classical data where their classical counterpart already established as the common way to solve the problem. An example for this approach is be the application of a quantum neural network on an image recognition task~\cite{beaulieu2022kernel-image-manufacturing-defects}. Our work definitely belongs to the latter group of articles and is to best of our knowledge the first one in the domain of PPM.

During our experiments, the applications of QML developed rapidly. A number of similar approaches explore QML algorithms in domains where data follow complex patterns. These include quantum kernel methods for high energy physics~\cite{wu2021kernel-lhc} or in aircraft design~\cite{yuan2022qsvm-aerodynamic} as well as in image recognition for manufacturing defects~\cite{beaulieu2022kernel-image-manufacturing-defects}. Classically, convolutional neural networks achieve high accuracy on image recognition tasks, which is why~\cite{matic2022quantum-conv-nn-radiology} investigate the advantage of quantum neural networks. On a more theoretical side, the work by~\cite{glick2022group-structure} shows experiments on synthetic data while ~\cite{huang2021quantum-data} and~\cite{jerbi2022qml-beyond} work with the MNIST-fashion dataset~\cite{xiao2017fashion-mnist}. Since the number of papers in QML is exploding over the last three years, there is a need for a more comprehensive review of recent applications.

The biggest issue with quantum computing is, that nowadays quantum hardware is in its very early phase of development~\cite{preskill2018nisq}. There is still a need for more high quality research on the theoretical side of the algorithms such as in~\cite{holmes2022untrainability}. Also, one has to see the full pipeline from hardware to end user and build applications that provide added value as a whole. To show, how a practical implementation of a QML pipeline to real hardware could look like, we further implemented an interface to the workflow engine Camunda in the form of an extension to the PPM plugin presented by~\cite{bartmann2021camunda-hpo}. The inter-case classifiers are implemented as a wrapper that connects to the Python evaluation framework and the quantum algorithms. It is then possible to connect a real IBM quantum computer hosted in the cloud. We believe that possible advantages and grievous limitations of current NISQ systems will become more aware when quantum technology finds its way out of academia and establishes in real-world applications. We hope that this work could show how quantum technology can be used in the process sciences, thereby serving as a draught horse for such applications~\cite{biamonte2017qml}. Those applications are not restricted to belong to the machine learning domain, but can also contain SAT solvers to find inconsistencies in business processes or apply the quantum approximate optimization algorithm (QAOA) to optimization such as assigning resource priorities to tasks.

\begin{figure}[ht]
	\centering
	\includegraphics[width=\textwidth]{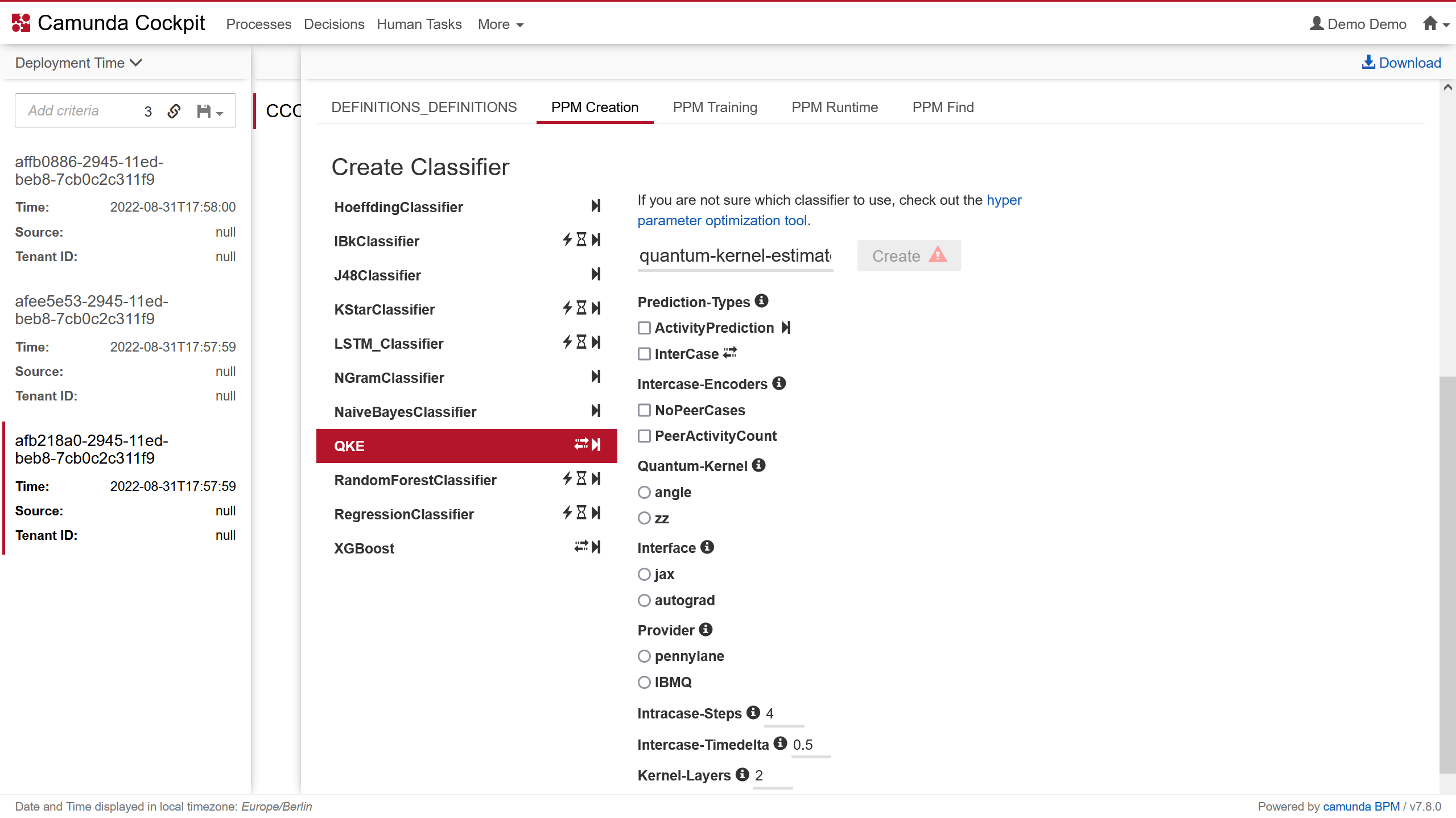}
	\caption{Quantum kernel estimator in the Camunda Cockpit during initialization.}
	\label{fig:qke-cockpit}
\end{figure}


\section{Conclusion}
In this paper, we proposed implicit quantum kernel methods to solve the inter-case PPM problem. Overall, the experiments showed, that quantum kernels achieve a practical improvement in accuracy compared to similar classical methods. While the robust quantum hardware still needs to scale, a first prototypical implementation proves that integration into existing workflow engines is feasible. For the intermediate time, we have shown empirically that undersampling to a certain amount leads to faster simulations of the quantum kernels without loss of accuracy. Conclusively, supported by our findings and a working demonstrator, we confidently call for courage to conduct further exploratory investigations on quantum algorithms in process sciences. Because BPM is an always evolving, but already a mature field of research, we claim it is the ideal candidate to take a leading role paving the way towards a new family of reliable quantum business applications.

\bibliographystyle{splncs04}
\bibliography{main}

\begin{thebibliography}{10}
\providecommand{\url}[1]{\texttt{#1}}
\providecommand{\urlprefix}{URL }
\providecommand{\doi}[1]{https://doi.org/#1}

\bibitem{arute2019google-sycamore}
Arute, F., {et al.}: Quantum supremacy using a programmable superconducting
  processor. Nature  \textbf{574}(7779),  505--510 (2019)

\bibitem{bartmann2021camunda-hpo}
Bartmann, N., Hill, S., Corea, C., Drodt, C., Delfmann, P.: Applied predictive
  process monitoring and hyper parameter optimization in camunda. In:
  Intelligent Information Systems, Lecture Notes in Business Information
  Processing, vol.~424, pp. 129--136. {Springer International Publishing}, Cham
  (2021)

\bibitem{beaulieu2022kernel-image-manufacturing-defects}
Beaulieu, D., Miracle, D., Pham, A., Scherr, W.: Quantum kernel for image
  classification of real world manufacturing defects. arXiv p. arXiv:2212.08693
  (2022)

\bibitem{biamonte2017qml}
Biamonte, J., Wittek, P., Pancotti, N., Rebentrost, P., Wiebe, N., Lloyd, S.:
  Quantum machine learning. Nature  \textbf{549}(7671),  195--202 (2017)

\bibitem{denisov2019performance-spectrum}
Denisov, V., Fahland, D., {van der Aalst}, W.M.: Predictive performance
  monitoring of material handling systems using the performance spectrum. In:
  2019 International Conference on Process Mining (ICPM). pp. 137--144. IEEE
  (2019)

\bibitem{difrancescomarino2022ppm}
{Di Francescomarino}, C., Ghidini, C.: Predictive process monitoring. In:
  Process Mining Handbook, Lecture Notes in Business Information Processing,
  vol.~448, pp. 320--346. {Springer International Publishing}, Cham (2022)

\bibitem{Egger2020QuantumComputingFinance}
Egger, D.J., Gambella, C., Marecek, J., McFaddin, S., Mevissen, M., Raymond,
  R., Simonetto, A., Woerner, S., Yndurain, E.: Quantum comp. for finance: Sota
  and future prospects. IEEE Transactions on Quantum Engineering  \textbf{1},
  1--24 (2020)

\bibitem{Emmanoulopoulos2022QuantumMachineLearning}
Emmanoulopoulos, D., Dimoska, S.: Quantum machine learning in finance: Time
  series forecasting. Advances in Computational Intelligence. MICAI 2022.
  Lecture Notes in Computer Science  (2022)

\bibitem{ferrini2021ln-qc}
Ferrini, G., {Frisk Kockum}, A., Vikst{\aa}l, P.: Lecture notes in quantum
  computing

\bibitem{feynman1982quantum-physics}
Feynman, R.P.: Simulating physics with computers. International Journal of
  Theoretical Physics  \textbf{21}(6-7),  467--488 (1982)

\bibitem{glick2022group-structure}
Glick, J.R., Gujarati, T.P., Corcoles, A.D., Kim, Y., Kandala, A., Gambetta,
  J.M., Temme, K.: Covariant quantum kernels for data with group structure

\bibitem{glosser2022bloch-sphere}
Glosser: Bloch sphere (2022), \url{https://bit.ly/bloch-sphere-wikipedia}

\bibitem{grinvald2021inter-case-variants}
Grinvald, A., Soffer, P., Mokryn, O.: Inter-case properties and process variant
  considerations in time prediction: A conceptual framework. In: Enterprise,
  Business-Process and Information Systems Modeling, Lecture Notes in Business
  Information Processing, vol.~421, pp. 96--111. {Springer International
  Publishing}, Cham (2021)

\bibitem{havlicek2019quantum-feature-space}
Havl{\'i}{\v{c}}ek, V., C{\'o}rcoles, A.D., Temme, K., Harrow, A.W., Kandala,
  A., Chow, J.M., Gambetta, J.M.: Supervised learning with quantum-enhanced
  feature spaces. Nature  \textbf{567}(7747),  209--212 (2019)

\bibitem{hey1999introduction-qc}
Hey, T.: Quantum computing: an introduction. Computing {\&} Control Engineering
  Journal  \textbf{10}(3),  105--112 (1999)

\bibitem{holmes2022untrainability}
Holmes, Z., Sharma, K., Cerezo, M., Coles, P.J.: Connecting ansatz
  expressibility to gradient magnitudes and barren plateaus. PRX Quantum
  \textbf{3}(1) (2022)

\bibitem{huang2021quantum-data}
Huang, H.Y., Broughton, M., Mohseni, M., Babbush, R., Boixo, S., Neven, H.,
  McClean, J.R.: Power of data in quantum ml. Nature communications
  \textbf{12}(1) (2021)

\bibitem{jerbi2022qml-beyond}
Jerbi, S., Fiderer, L.J., Nautrup, H.P., K{\"u}bler, J.M., Briegel, H.J.,
  Dunjko, V.: Quantum machine learning beyond kernel methods

\bibitem{Kandala2017HardwareefficientVariationalQuantum}
Kandala, A., Mezzacapo, A., Temme, K., Takita, M., Brink, M., Chow, J.M.,
  Gambetta, J.M.: Hardware-efficient variational quantum eigensolver for small
  molecules and quantum magnets. Nature  \textbf{549}(7671),  242--246 (2017)

\bibitem{klijn2020performance-spectrum}
Klijn, E.L., Fahland, D.: Identifying and reducing errors in remaining time
  prediction due to inter-case dynamics. In: 2020 2nd International Conference
  on Process Mining (ICPM). pp. 25--32. IEEE (2020)

\bibitem{leontjeva2015complex}
Leontjeva, A., Conforti, R., {Di Francescomarino}, C., Dumas, M., Maggi, F.M.:
  Complex symbolic sequence encodings for predictive monitoring of business
  processes. In: Business Process Management, Lecture Notes in Computer
  Science, vol.~9253, pp. 297--313. {Springer International Publishing}, Cham
  (2015)

\bibitem{maggi2014predictive-process-monitoring}
Maggi, F.M., {Di Francescomarino}, C., Dumas, M., Ghidini, C.: Predictive
  monitoring of business processes. In: Advanced information systems
  engineering, Lecture Notes in Computer Science, vol.~8484, pp. 457--472.
  Springer, Cham (2014)

\bibitem{matic2022quantum-conv-nn-radiology}
Matic, A., Monnet, M., Lorenz, J.M., Schachtner, B., Messerer, T.:
  Quantum-classical convolutional neural networks in radiological image
  classification

\bibitem{Pistoia2021QuantumMachineLearning}
Pistoia, M., Ahmad, S.F., Ajagekar, A., Buts, A., Chakrabarti, S., Herman, D.,
  Hu, S., Jena, A., Minssen, P., Niroula, P., Rattew, A., Sun, Y., Yalovetzky,
  R.: Quantum machine learning for finance. IEEE/ACM International Conference
  On Computer Aided Design (ICCAD), Munich, Germany  (2021)

\bibitem{pourbafrani2022inter-case-remaining-time}
Pourbafrani, M., Kar, S., Kaiser, S., {van der Aalst}, W.M.P.: Remaining time
  prediction for processes with inter-case dynamics. In: Process Mining
  Workshops, Lecture Notes in Business Information Processing, vol.~433, pp.
  140--153. {Springer International Publishing}, Cham (2022)

\bibitem{preskill2018nisq}
Preskill, J.: Quantum computing in the nisq era and beyond. Quantum
  \textbf{2}, ~79 (2018)

\bibitem{rama-maneiro2022deep-learning-benchmark}
Rama-Maneiro, E., Vidal, J., Lama, M.: Deep learning for predictive business
  process monitoring. IEEE Transactions on Services Computing p.~1 (2022)

\bibitem{scholkopf2001representer}
Sch{\"o}lkopf, B., Herbrich, R., Smola, A.J.: A generalized representer
  theorem. In: Computational Learning Theory, Lecture Notes in Computer
  Science, vol.~2111, pp. 416--426. {Springer Berlin Heidelberg}, Berlin,
  Heidelberg (2001)

\bibitem{scholkopf2002kernel}
Sch{\"o}lkopf, B., Smola, A.J.: Learning with Kernels: Support vector machines,
  regularization, optimization, and beyond. Adaptive computation and machine
  learning series, {MIT Press}, Cambridge, Mass. (2002)

\bibitem{schuld2020circuit-centric}
Schuld, M., Bocharov, A., Svore, K.M., Wiebe, N.: Circuit-centric quantum
  classifiers. Physical Review A  \textbf{101}(3) (2020)

\bibitem{schuld2019feature-space-extended}
Schuld, M., Killoran, N.: Quantum machine learning in feature hilbert spaces.
  Physical review letters  \textbf{122}(4),  040504 (2019)

\bibitem{schuld2020encoding}
Schuld, M., Sweke, R., Meyer, J.J.: Effect of data encoding on the expressive
  power of variational quantum-machine-learning models. Physical Review A
  \textbf{103}(3) (2021)

\bibitem{senderovich2017intra-inter}
Senderovich, A., {Di Francescomarino}, C., Ghidini, C., Jorbina, K., Maggi,
  F.M.: Intra and inter-case features in predictive process monitoring: A tale
  of two dimensions. In: Business Process Management, Lecture Notes in Computer
  Science, vol. 10445, pp. 306--323. {Springer International Publishing}, Cham
  (2017)

\bibitem{senderovich2019inter-case-encoding}
Senderovich, A., {Di Francescomarino}, C., Maggi, F.M.: From knowledge-driven
  to data-driven inter-case feature encoding in predictive process monitoring.
  Information Systems  \textbf{84},  255--264 (2019)

\bibitem{shor1994primes}
Shor, P.W.: Algorithms for quantum computation: discrete logarithms and
  factoring. In: Proceedings 35th Annual Symposium on Foundations of Computer
  Science. pp. 124--134. {IEEE Comput. Soc. Press} (1994)

\bibitem{Streif2021Beatingclassicalheuristics}
Streif, M., Yarkoni, S., Skolik, A., Neukart, F., Leib, M.: Beating classical
  heuristics for the binary paint shop problem with the quantum approximate
  optimization algorithm. Physical Review A  \textbf{104}(1),  012403 (2021)

\bibitem{teinemaa2019diss}
Teinemaa, I.: Predictive and Prescriptive Monitoring of Business Process
  Outcomes. {University of Tartu Press}, Tartu (2019)

\bibitem{teinemaa2019benchmark-outcome-oriented}
Teinemaa, I., Dumas, M., {La Rosa}, M., Maggi, F.M.: Outcome-oriented ppm. ACM
  Transactions on Knowledge Discovery from Data  \textbf{13}(2),  1--57 (2019)

\bibitem{vapnik1995svm}
Vapnik, V.N.: The Nature of Statistical Learning Theory. Springer eBook
  Collection Mathematics and Statistics, Springer, New York, NY (1995)

\bibitem{wu2021kernel-lhc}
Wu, S.L., {et al.}: Application of quantum ml using the quantum kernel
  algorithm on high energy physics analysis at the lhc. Physical Review
  Research  \textbf{3}(3) (2021)

\bibitem{xiao2017fashion-mnist}
Xiao, H., Rasul, K., Vollgraf, R.: Fashion-mnist: a novel image dataset for
  benchmarking machine learning algorithms. arXiv e-prints p. arXiv:1708.07747
  (2017)

\bibitem{Yamamoto2019naturalgradientvariational}
Yamamoto, N.: On the natural gradient for variational quantum eigensolver.
  arXiv:1909.05074 [quant-ph]  (2019)

\bibitem{yuan2022qsvm-aerodynamic}
Yuan, X.J., Chen, Z.Q., Liu, Y.D., Xie, Z., Jin, X.M., Liu, Y.Z., Wen, X.,
  Tang, H.: Quantum support vector machines for aerodynamic classification.
  arXiv e-prints p. arXiv:2208.07138 (2022)

\bibitem{zhong2020quantum-supremacy}
Zhong, H.S., {et al.}: Quantum computational advantage using photons. Science
  (New York, N.Y.)  \textbf{370}(6523),  1460--1463 (2020)

\end{thebibliography}

\end{document}